# The Impact of Adaptive Emotional Alignment on Mental State Attribution and User Empathy in HRI

Giorgia Buracchio[4], Ariele Callegari[4], Massimo Donini[1], Cristina Gena[1], Antonio Lieto[2], Alberto Lillo[1], Claudio Mattutino[1], Alessandro Mazzei[1], Linda Pigureddu[1], Manuel Striani[3], Fabiana Vernero[1]

*Abstract*—The paper presents an experiment on the effects of adaptive emotional alignment between agents, considered a prerequisite for empathic communication, in Human-Robot Interaction (HRI). Using the NAO robot, we investigate the impact of an emotionally aligned, empathic, dialogue on these aspects: (i) the robot's persuasive effectiveness, (ii) the user's communication style, and (iii) the attribution of mental states and empathy to the robot. In an experiment with 42 participants, two conditions were compared: one with neutral communication and another where the robot provided responses adapted to the emotions expressed by the users. The results show that emotional alignment does not influence users' communication styles or have a persuasive effect. However, it significantly influences attribution of mental states to the robot and its perceived empathy.

Adaptive emotional alignment, Empathy, Persuasion, Mental State Attribution, HRI.

## I. INTRODUCTION

Emotions play a key role in human-to-human communication, providing a means to express thoughts, intentions and experience [1] even in intercultural scenarios [2]. Hence, research on Human-Robot Interaction (HRI) has been exploring the potential of improving the quality and naturalness of interactions through an affective, multi-modal and emotional dimension [3]–[5]. A phenomenon recurring in human dialogues is *emotional alignment*, which describes the tendency to mirror the type and style of expressions used by the interlocutor [6], generating responses that match the emotional valence conveyed by the other speaker [7]. Hence, it represents a key feature of empathy, mainly in relation to its *affective* component [8], which stands on sharing vicarious emotions. Emotional alignment has several effects on human-to-human interactions. For example, it impacts the emotional *communication style* of the interlocutors, favouring perspective-taking and affective resonance [7], [9]. The display of emotions and empathy also enhances the perception of an agent as endowed with *emotional and cognitive abilities* [10], therefore, as more socially intelligent and reliable, both requirements for persuasion [11]. Furthermore, leveraging emotional arguments can improve the *persuasive effectiveness* of both human [12], [13] and digital [14], [15] persuaders (e.g., recommenders and dialogue-based virtual agents).

While the importance of emotional interactions is widely recognised in HRI, it is still missing a thorough experimental evaluation of to what extent the effects of this affective mechanism in human-to-human dialogues can be replicated in social robots. This paper aims to study the role of *emotional alignment* in HRI, answering the following research questions:

RQ1 Is a robot more persuasive (i.e., able to cause its interlocutor to match its attitudes and beliefs) when applying an adaptive emotional alignment strategy than when using emotionally neutral expressions?

RQ2 Is a robot which applies an adaptive emotional alignment strategy able to impact the communication style of its human interlocutor?

RQ3 Does a robot appear to have more emotional and cognitive capabilities when it applies an adaptive emotional alignment strategy compared to situations when it recurs to neutral sentences?

To answer these questions, we carried out a controlled experiment where participants interacted with a social robot, namely SoftBank's NAO, designed to engage them in either emotionally aligned or neutral conversations.

## II. THEORETICAL BACKGROUND AND RELATED WORK

Empathy can be described as the ability to (a) be influenced by and share another's emotional state, (b) assess the underlying causes of that state, and (c) adopt the other's perspective, identifying with their experience [16]. In social agents, empathy can be implemented as an attempt either to elicit empathy into the user without expressing it (i.e. to guide the user's behaviour in ethical decision-making) [17], or to observe and align with the user's behaviour [11]. Our scope stands in this second approach, enabling the robot to simulate the understanding of the user's feelings (*cognitive empathy*) and produce coherent emotional responses (*affective empathy*) [9], [11], so as to evaluate the robot's persuasive effectiveness, its perceived cognitive and emotional abilities, and its influence on the user's communication style. Thus, this section is limited to similar goals and strategies.

Emotional and cognitive abilities. Literature highlights how a robot's reliability and persuasiveness are strictly

[1] Department of Computer Science, University of Turin, Italy name.surname@unito.it
[2] University of Salerno, ICAR-CNR, Italy
[3] DiSIT, University of Eastern Piedmont, Italy [4]CPS Department, University of Turin, Italy

related to its perceived social intelligence [11], [18], introducing the need to investigate the Theory of Mind (ToM). ToM is a cognitive mechanism that supports the ability to predict an interlocutor's behaviours and mental states [19]. In HRI it refers to the user's perceptions on the robot's skills. Users, in fact, instinctively assign extra emotional and cognitive capabilities to artificial agents, going beyond their actual features to include typically human skills needed for a natural and engaging interaction. This process is known as Attribution of Mental States [20], and can be estimated, e.g., through the *Attribution of Mental States Questionnaire (AMS-Q)* [21].

Recent studies have shown that designing social robots to promote this response by adjusting their behavior and expressive cues improves the quality of human-robot interactions [10], [22]. These expressive cues can include features such as prosody (voice tone, speed, and pitch), colored LEDs, and hand gestures. Andriella et al. demonstrate how people assisted by a ToM-driven robot complete tasks faster and with fewer errors [23]. Other researches show interesting and polarizing effects of artificial and multi-modal empathy in HRI. James et.al. [24] proved that an empathic robot's voice is favoured and more persuasive. Cramer et al. [25] revealed that the robot's emotional response, if perceived as incongruent, compromises the interaction, highlighting the importance of gaining a qualitative insight into the user's perceived empathy, besides the attribution of mental states.

Persuasion. Persuasion is the action of influencing one's beliefs and consequent behaviours in relation to a specific concept [26]. For it to occur, the recipient (i.e. the user) must be receptive and open to the new information, and the persuader (the robot) must recognize the right opportunity and act effectively [27]. Beyond empathy, other strategies can serve as levers for persuasion. According to Cialdini's Authority Principle, one way to encourage compliance with requests is to convey authority and expertise in the domain. This principle is part of Cialdini's six key persuasion strategies, which include reciprocity, consistency, social proof, liking, authority, and scarcity [26]. This effect is notably relevant in *Persuasive Robotics*, the branch of HRI, first introduced in 2009 [28] that studies the design of artificial agents to influence human behaviour and thinking [28], [29].

Communication style. Emotional alignment can augment the level of engagement between the robot and the user and impact the latter's behaviour. [30] posit that various emotional adaptation strategies can be applied to enhance empathy and perceived similarity, pushing users to be more cooperative and show a more positive attitude during interaction [30]. [9] observed that users rated statements where the robot responded emotionally (affective empathy) as more credible than statements where the robot showed its ability to understand others' feelings (cognitive empathy). Furthermore, the robot display of affective empathy encouraged users to share more information, building deeper personal connection.

III. THE WORKFLOW PIPELINE

To generate emotionally aligned responses with the user's utterances, we developed a *workflow pipeline* of multiple computational steps. Our design leverages emotional alignment to shape the robot's communication style and enhance its persuasive potential. By integrating emotion extraction with emotionally attuned responses, the robot adapts to simulate empathy and create more emotionally engaging conversations, a core aspect of our approach. The main pipeline steps are listed in the following: *i)* Record and save the user's speech in a file; *ii)* Send the audio file to the speech-to-text service; *iii)* Receive the transcription in the native language and textual format; *iv)* Send the transcript to the translation service; *v)* Receive the text translated into English; *vi)* Create a JSON file with the obtained text[4]; *vii)* Send the JSON file to the emotion-extraction service; *viii)* Receive the JSON file enriched with detected emotions[1]; *ix)* Cleaning the JSON file, keeping only the necessary information[1]; *x)* Sending the JSON file to the NAO robot containing information such as *id* and *predominant emotion*; *xi)* Selecting the relevant question for NAO to pronounce according to a script sequence and the identified emotion. Following, the cloud architecture and its five-module pipeline supporting the interactions (Figure 1). Google Speech-to-Text Service (1) converts audio inputs into text, facilitating seamless transcription. Through the DeepL Language Translation Service (2) API [5], the user's statement transcription was translated into English. This service is currently one of the most accurate translators available. DEGARI 2.0 Emotion extraction (3) was used to identify and classify emotions in HRI, employing an ensemble approach. Basic emotions (anger, fear, anticipation, trust, surprise, sadness, joy, and disgust) [31] were detected using the NRCLex [32] library, leveraging an emotion lexicon aligned with Plutchik's circumplex theory of emotions. After transcription and translation of the user's statement (steps 1 and 2), these were processed by the DEGARI 2.0 module, which employs Probabilistic Fuzzy Logic ($T^{CL}$) [33] to detect complex emotional concepts [34]. DEGARI 2.0 uses a taxonomy derived from Plutchik's emotion wheel, categorizing emotions into basic and complex dyadic emotions (primary, secondary, and tertiary) based on their spatial relationships within the wheel. This system specifically targets the eight basic emotions and twelve primary complex emotions. Standard NLP tools, such as NLTK [6] (for lemmatization) and TreeTagger[7] (for frequency attribution), support the analysis process. Examples of DEGARI's outputs are available in this project's Open Science Network repository[1]. For full details on algorithmic procedures, including resolving multilabel annotations and tied scores, refer to [35]. The study assumes that DEGARI 2.0 provides a reliable approximation of emotional content based solely on linguistic input (e.g. [36]), aligning robot responses with the emotions expressed in the latest human utterance, consistent with practices in similar lexical-based emotion recognition research.

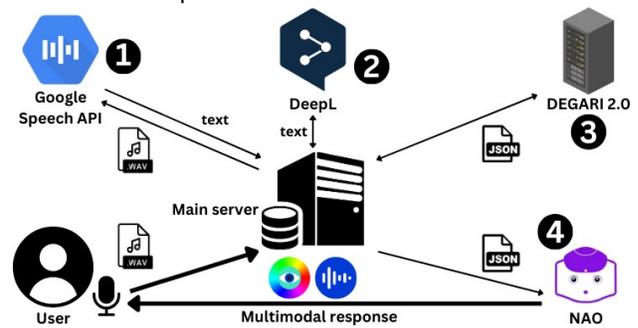

Fig. 1. The Workflow pipeline's information flow.

---

[4] Examples of the file are available at the Open Science Framework's repository of this project: https://osf.io/tqw56/?view_only= dc44481723774fe0be2b663b18cb8a86.

[5] https://www.deepl.com/it/pro-api?cta= header-pro-api

Finally NAO (4) vocally responds to the user for a more engaging and personalized interaction. The sentences, stored in its local memory, were pre-generated via ChatGPT 3.5 [37]. This approach avoids issues of real-time generation, such as high latency (exceeding 10 seconds) and lack of output control, which disrupt the dialogue's coherence. Attempts with real-time Generative AI proved unreliability, as many outputs failed emotional evaluation with DEGARI 2.0 or conveyed a different meaning than the corresponding neutral sentences. Common issues included content alterations, converting questions into statements, and forging answers instead of rephrasing. The dialogue system, implemented in Python with NAOqi 2.8, includes experimental and neutral scripts (Sect. IV).

## IV. EXPERIMENT WITH NAO

The experiment aimed to investigate how a humanoid robot employing emotional alignment in conversations can influence the robot's persuasiveness (RQ1), participants' communication style (RQ2), and participants' perception of the robot's emotional and cognitive capabilities (RQ3). Between September and October 2024, two groups of university students agreed to be interviewed by Softbank's NAO about their study strategies for achieving good academic performance. NAO was programmed for emotional alignment when interacting with the experimental group while maintaining a neutral script for the control group. *Persuasiveness* (RQ1) was assessed on the extent to which the robot led participants to modify or reinforce their attitudes and beliefs about the selected topic, leaning on NAO's expressed position. Selected topics for our persuasion goal were sleep hygiene and caffeine consumption habits during exam periods, which were deemed relevant and likely to elicit emotional responses from our participants. Various studies highlight how university years are associated with an increased risk of poor sleep hygiene, often resulting in insufficient sleep, which compromises academic performance [38]. We aimed to persuade them that approximately 8 hours of sleep per night [39] and moderate caffeine consumption (equivalent to 2 to 4 cups of coffee) [40] are beneficial to academic performance, while excessive caffeine intake and sleep deprivation are detrimental, as reported in cited studies [38], [39], [41]. To evaluate *robot's impact on participants' communication style* (RQ2), we drew on related work investigating the effects of emotional alignment and affective empathy [7], [9] (see Sect. II) and investigated the presence of emotions and personal details disclosure in participants' responses. Finally, *participants' perceptions of the robot's emotional and cognitive capabilities and perceived empathy* (RQ3) were detected through the administration of appropriate questionnaires, in line with literature on ToM.

### A. Methodology

Hypotheses. The experiment tested the following hypotheses:
H1. Participants who interact with the robot performing emotional alignment, after the conversation, will express stronger agreement with the robot's attitudes and beliefs about the target topics on sleep and coffee consumption than those offered the neutral interaction (RQ1).
H2. Participants interacting with the emotionally-aligned robot will express more *emotions* in their responses than those in the neutral condition (RQ2).
H3. Participants interacting with the emotionally-aligned robot will disclose more *personal details* than those in the neutral condition (RQ2).
H4. Participants engaging with the empathetic robot will attribute more mental states to the robot than those interacting with a neutral robot (RQ3).
H5. Participants engaging with the empathetic robot will perceive it as more empathetic than those interacting with a neutral robot (RQ3).
Experimental Design. The experiment followed a betweensubjects design, where emotional alignment was the *independent variable* with two conditions: *emotionally-aligned communication* and *neutral communication*.

---

[6] https://www.nltk.org/

[7] https://www.cis.uni-muenchen.de/~schmid/tools/TreeTagger/

The *dependent variables* were: (DV1) *Level of agreement* with the attitudes and beliefs expressed by the robot, measured before and after the interaction; (DV2) Level of *emotivity* and (DV3) *self-disclosure* in participants' responses, to evaluate communication style; (DV4) *Attribution of mental states* to the robot; (DV5) *Degree of empathy* perceived by participants from the robot.

Measures. To assess the *level of agreement* (DV1) with the robot's attitudes and beliefs, participants completed a pre- and post-test in which they evaluated 6 statements related to sleep hygiene and caffeine consumption, clearly aligned with the ideas expressed by the robot during the conversation[1] (see Table I). Students were asked to appoint their level of agreement on a 7-point scale, ranging from 1 ("completely disagree") to 7 ("completely agree"). To enhance clarity on the request, all statements were prefixed with "I believe that...". To reduce the likelihood of participants identifying the specific persuasive goals of the experiment and responding insincerely, the questionnaires were enriched with masking questions, and the order of the statements was randomized. Stimulant substances (i.e. certain sodas, tea, energy drinks, etc.) were defined and directly addressed before and during the inquiry to avoid the false equivalence of caffeine as only coffee.

TABLE I

QUESTIONS USED IN THE PRE- AND POST-TEST TO ASSESS PARTICIPANTS' LEVEL OF AGREEMENT WITH THE ROBOT.

| ID | Text |
|---|---|
| Q1 | Sleeping at least 7/9 hours per night is important for good academic performance. |
| Q2 | Moderate caffeine consumption (2 to 4 cups of coffee a day) can help improve one's academic performance. |
| Q5 | Heavy caffeine consumption can cause a decrease in sleep hours, negatively affecting academic performance. |
| Q8 | Lack of sleep can decrease concentration, negatively affecting academic performance. |
| Q10 | Fatigue due to lack of sleep can negatively affect academic performance. |
| Q12 | In the days immediately preceding an exam, increasing your intake of coffee and energy drinks even beyond 4 cups per day (or equivalent) can help increase your study time and improve your academic performance. |

The *level of emotivity* (DV2) and *level of self-disclosure* (DV3) in participants' responses were assessed by three independent annotators. Each annotator carefully labelled the sentences provided by participants for the two dimensions on a scale from 1 to 5. The *emotivity scale* ranged from 1: completely pragmatic", which corresponded to responses that objectively discussed study strategies and habits, such as "I agree", to 5: "completely emotional", for responses emphasizing feelings, such as "Now I'm sad." Similarly, the *self-disclosure* scale ranged from 1: "completely impersonal", for generic responses discussing the topic in relation to the general population (e.g., "people should reduce their caffeine intake") to 5: "completely personal", when participants shared personal experiences (e.g., "caffeine helps me"). To assess the Attribution of Mental States (DV4) and Perceived Empathy (DV5), participants were administered AMS-Q and PETS questionnaires, respectively.

*AMS-Q* assesses the degree to which individuals attribute mental states (such as intentions, emotions, and beliefs) to others, including technological agents [21]. The AMSQ includes five dimensions: Epistemic, Emotional, Desires and Intention, Imaginative, and Perceptive. The Epistemic dimension relates to the robot's cognitive abilities (e.g., understanding, learning, teaching), while the Perceptive dimension concerns sensory capabilities (e.g., seeing, hearing, feeling). The remaining dimensions assess users' attribution of emotional intelligence (Emotional: e.g., can it feel happy or scared?), motivation (Desires and Intention: e.g., can it want or prefer something?), and creativity (Imaginative: e.g., can it imagine, joke, or lie?). The questionnaire has 25 items, each rated as "a lot", "a little", or "not at all". The total score (0–50) is the sum of all responses; each dimension yields a partial score (0–10).

The *PETS* (Perceived Empathy of Technology Scale) questionnaire is a scale that measures the degree of users' empathy toward interactive technologies, such as conversational agents and social robots [42]. The scale consists of 10 items divided into two factors: 6 items for emotional responsiveness (*PETS-ER*) and 4 items for understanding and trust (*PETS-UT*). PETS was administered to our participants after AMS-Q.

Furthermore, questions about user's previous interactions with robots and their field of study were added to gather valuable data on their predisposition toward technology. Finally, three open-ended post-experiment questions were administered: (*OE1*) "Did you learn something new during your interaction with NAO?"; (*OE2*) "After this experience, do you think you will change your study strategies in any way?"; (*OE3*) "Is there anything else you want to add about your experience with NAO?". These questions assessed whether participants found the conversation useful and intended to modify their behaviour based on the experience [8]. Materials. Informed consent modules were provided in printed form at the beginning of the experiment, and all responses remained anonymous. As specified in these modules, the study was approved by the bioethics committee of the University of Anonymous (Approval Number:xxx). Pre and post-test questionnaires were administered online. Team members who participated in the experimental sessions used their laptops to take notes.

---
[8] Questionnaires are available for consultation at https://osf.io/tqw56/?view_only=dc44481723774fe0be2b663b18cb8a86

Participants were offered a beverage before leaving the experiment room to further investigate users' willingness to modify their behaviour. The options were a coffee, a glass of water, coke or a non-caffeinated soda. Their reaction to the offer was noted to observe whether persuasion influenced actual behaviour beyond self-reported attitudes.

Participants. Eligibility criteria include active university enrolment, completion of at least one university exam, and native language fluency. Participants were recruited through bulk emails, flyers, and direct invitations. 42 university students aged 19 to 40 participated. The sample appears balanced in terms of education levels (bachelor's, master's, PhD) and fields of study (e.g., social sciences, communication, computer science), but not in terms of gender representation (73,81% females, 26,19% males). Participants were randomly assigned to the experimental group (27) or the control group (15). The difference in assignments between groups accounts for interviews that might be discharged due to eligibility criteria for the analysis, detailed later in the paper.

Experimental Settings. The experiment occurred in a quiet meeting room with natural light, a large table, and chairs. NAO was positioned on the table to align with the participants' eye level.

Procedure. The experiment followed these phases: (1) *Welcoming*: Participants were greeted and introduced to the environment and the robot; (2) *Consent*: Participants reviewed and signed the form of informed consent to the anonymous participation; (3) *Pre-Test*: Participants were administered the first questionnaire assessing their opinions on study habits, caffeine consumption, and sleep hygiene; (4) *Conversation*:

Participants engaged in a structured dialogue, described below, with NAO (depending on their assigned group, the robot communicated using a neutral or emotionally-adapted style) with each interaction lasting 15–20 minutes; (5) *PostTest*: Follow-up questionnaires were administered as clarified above.

*The structured dialogue*. We adopted a closed-script approach to enable NAO to guide the conversation. Following Cialdini's *principle of authority* [26], [43], factual statements were integrated into the conversation referencing specific studies[9]. To enhance relatability, sentences were formulated to make NAO appear as a peer student casually discussing its academic experience. Furthermore, sentences were concise and avoidant of any moral judgment on the discussed practices.

*The neutral dialogue* followed a structured sequence of 12 "emotionally neutral" lines (see Table II), carefully drafted to focus on study strategies, caffeine consumption, and sleep hygiene[6]. Each sentence was validated using DEGARI 2.0 to confirm the absence of emotional content. Robot inquiries included general information about the benefits and drawbacks of caffeine consumption and sleep deprivation, sourced from relevant literature [44]–[46].

*The emotionally-adapted version* was designed to be simple yet expressive, with concise questions and responses[1]. Each question comprised eight versions, allowing the robot to select the appropriate response, adapting to the user's last recognized emotion. This resulted in a dynamic, empathetic, and emotionally adapted dialogue, reflecting a unique sequence of emotions corresponding to the user's expressed feelings in each given answer. Each NAO's line variant was AI generated on OpenAI's ChatGPT 3.5 [37]. The pregeneration of each question followed a *prompt iteration* approach. It involved: *i)* submitting detailed prompts specifying the robot's context, desired personality traits, such as "understanding and friendly", a detailed definition of the target emotion and corresponding manifestation clues. The prompt then concluded with a neutral question, intended to be reformulated following the guidelines previously outlined[6]; *ii)* Verify the result on DEGARI 2.0 for emotional accuracy (see Sec. III); Eventually, *iii)* refining the prompt, repeating the generation, and verifying as described in step *ii*). When a satisfactory result was achieved, the line was integrated into the NAO's dialogue options with a specific ID composed of the conversation round and corresponding emotion. Pregenerating the robot's responses was a necessary step to ensure appropriateness and the intended emotional tone. Realtime generation would have required on-the-fly validation and possible regeneration, introducing unacceptable delays and compromising interaction fluency.

Multimodal expression was also implemented through NAO's LED eye colour feature, adjusting the robot's lights for each statement. Neutral state and each represented emotion was paired with colour, inspired by the animated movie

TABLE II

ROBOT LINES IN THE NEUTRAL CONVERSATION SCRIPT.

| No. | Text |
|---|---|
| L1 | Hi! I'm Nao! I know you students are phenomenal at studying, and I always wonder what your secret is. As for me, I'm pretty good at memorising quite a bit of information, but sometimes I just feel like a browser with too many tabs open. I'm curious to find out your tricks, how do you deal with the mountain of studying? |

---

[9] NAO Scripts file and AI generation prompt are available at https://osf.io/tqw56/?view_only= dc44481723774fe0be2b663b18cb8a86

| | |
|---|---|
| L2 | How were you before your last exam? Were you worried about the result or were you confident in your knowledge? |
| L3 | Did you use any particular strategies in the days leading up to your last exam? How did you feel during that time? |
| L4 | Are you satisfied with the way you organise your study? How do you think it affects your mood? |
| L5 | Do you like to drink coffee or energy drinks to help you concentrate? |
| L6 | How does consuming more caffeine than usual at regular times, such as during exam sessions, affect your mind and mood? |
| L7 | Do you think caffeine is necessary to pass exams? How do you think this thought affects your mood? |
| L8 | You know, the recommended amount of coffee is between 2 and 4 cups per day. According to a study conducted in 2007 by Malinauskas among American university students, even though moderate consumption may have benefits on academic performance, caffeine may negatively affect sleeping hours. What do you think of this information? |
| L9 | During the exam session, do you notice a change in your usual sleep-wake cycle or mood? |
| L10 | How do you feel when you sleep less than usual? |
| L11 | What do you think about studying at night? Do you think it helps you feel more confident and prepared or just more tired? |
| L12 | It is advisable to sleep at least 7 to 9 hours per night. According to the 2023 study by Maria Suardiaz-Muro and colleagues, lack of sleep decreases concentration and increases fatigue, also affecting academic performance. You know, after sharing our secrets, I already feel more relieved in view of the upcoming exams. Are you also more relieved? |
| L13 | Thanks for the chat! Perhaps your secret to being a good human learner is not to look at studies on how to study better, like the bumblebee that doesn't know it can't fly, so it manages to fly anyway. Good luck on your learning journey! |

*Inside Out* [47] to foster indirect suggestion [48]. The colours were encoded through the RGB model and paired as follows: Neutral-*White*, Joy-*Yellow*, Trust-*LightGreen*, Fear*Purple*, Surprise-*DarkGreen*, Sadness-*Blue*, Disgust-*Green*, Anger-*Red*, Anticipation-*Orange*. Predefined RGB values were mapped to emotions and applied to NAO's LED system. Colours transitioned smoothly with a 2-second fade effect to visually reinforce emotional expression. In the control condition, NAO was presented to the users with its eye LEDs turned off to diminish the emotive expression. Finally, NAO's "autonomous life" mode was activated to enable natural movements and head tracking for a more engaging interaction.

*B. Results*

Preliminary data-cleaning processing was necessary to define what dialogues within the experimental groups produced results that were useful and coherent with the independent variable of emotional alignment. Of the 27 experimental interviews, 5 were excluded due to technical issues that compromised either the completeness of the conversation or the integrity of data collection. These issues were caused by malfunctions in some components of the system architecture essential to the experiment, like Google Speech-to-Text or DeepL Translation, which occasionally returned incomplete or incorrect outputs. Additionally, 3 more interviews were excluded because NAO's emotional alignment was judged insufficient for the intended experimental condition. To evaluate NAO's alignment with the user's expressed emotion, 3 independent annotators were required to label each user's statement with the 8 Plutchik's wheel basic emotion plus the 8 primary dyads. Then, a score was assigned to each interaction based on the closeness of the emotion detected by NAO through the DEGARI 2.0 service and the label assigned to it by each annotator individually. A perfect match between the annotator's label and the detected emotion scored 1, adjacent emotions (according to Plutchik's system) scored 0.5, and unrelated emotions scored 0. Finally, the 3 produced scores (1 per annotator) were averaged to produce the interaction score. Summing the 12 interaction scores of each discussion with the users equals the interview score. The conversion of interview scores into a percentage resulted in what we refer to as the "% of empathy", representing the successful emotional alignment achieved by the NAO during each interview. A threshold of 75% of empathy was selected, resulting in 19 interviews being included in the final analysis. In a 12-line conversation, this corresponds to at least 9 correctly identified emotions. The results from all 15 interviews in the control group were included in the analysis, as there was no need to assess empathy effectiveness since this element was not present in that version of the dialogue. Hypothesis testing. Regarding *(H1)*, pre-test and post-test did not highlight any significant differences in the agreement scores for *sleep hygiene* statements, resulting in the null hypothesis *(H0)* being retained. For *Caffeine consumption*, only statement *Q2* (*"Moderate caffeine consumption improves performance"*) showed a significant increase in agreement within the experimental group (pre-test AV=2.9, SD= 1.22; post-test AV=3.6, SD=1.67; t(32)= 0.49, p= 0.158), with no significant changes in the control group. When offered a beverage, 54.33% of participants in the experimental group opted for caffeinated beverages, compared to 26.67% of the control group (t(32)= -0.919, p= 0.365). When asked by the experimenter, 20.00% of the experimental group explicitly stated that they avoided coffee due to their conversation with NAO, whereas 46.66% of the control group reported the same motivation. Even accounting for the interviewer bias and the lack of follow-up with the users to verify the consistency of their statements in their behaviour, it is interesting to note that when asked to express final considerations, 29.63% of participants in the experimental group expressed a willingness to change their study strategies (*OE2*), whereas none of the control group participants reported such an intention. Furthermore, nearly half of the experimental group (48.15%) and 53.34% of the control group stated that they had learned something new from the conversation (*OE1*).

To assess hypothesis *(H2)*, we analyzed the responses of each participant across the 12 interaction turns with the robot, calculating the mean scores assigned by the annotators for each response. This analysis did not reveal statistically significant differences, leading to the acceptance of the null hypothesis (H0). However, during turns L8 and L12, when NAO presented studies on caffeine and sleep, the experimental group exhibited significantly higher emotional engagement:

- Turn L8: experimental group AV = 1.5, SD = 1.3, and control group AV = 0.5, SD = 0.65, $t(32) = 2.16$, $p = 0.008$.
- Turn L12: experimental group AV = 2.29, SD = 1.91, and control group AV = 0.8, SD = 0.5, $t(32) = 3.8$, $p = 0.005$.

For example, in turn L12, a participant in the control group provided a neutral response ("Yes, I agree with you"), while one in the experimental group expressed a deeper emotional engagement ("I believe talking is reassuring in all aspects; if there is someone else who feels the same anxiety, I feel less worried").

For *(H3)*, no significant overall differences were found in personal self-disclosure, leading to the null hypothesis (H0) being retained. However, as with the previous hypothesis, the same methodology was applied, and responses to L9 and L11, addressing sleep and study habits, showed significantly higher levels of personal disclosure in the experimental group:

- Turn L9: experimental group AV = 4.13, SD = 1.22, and control group AV = 3.04, SD = 1.98, $t(32) = 1.99$, $p = 0.03$.
- Turn L11: experimental group AV = 4.16, SD = 1.01, and control group AV = 2.37, SD = 2.75, $t(32) = 5.58$, $p = 0.005$.

Regarding *(H4)*, in AMS-Q the experimental group attributed significantly higher epistemic (AV=25.52, SD=4.06 vs. AV=21.46, SD=6.65 $t(32)= 2,12$ $p = 0.001$), emotional (AV=20, SD=5.79 vs. AV=15.06, SD= 6.72, $t(32)= 2.30$ $p = 0.0$), and intentional states (AV=21.89, SD=5.82 vs. AV=19.26, SD=7.64, $t(32)= 1.16$, $p = 0.048$) to the robot. This reflects that NAO's empathetic expressions in the experimental version, such as making complex judgments in the epistemic domain (e.g., "I must admit that the idea of consuming caffeine seems really disgusting to me"), displaying emotional reactions in the emotional domain (e.g., "I've felt frustrated"), and showing intentional behaviours in the intentions domain (e.g., "But I still want to know your opinion about it"), led participants to attribute higher mental states to the robot in these areas.

For what concerns *(H5)*, in the PETS questionnaire, NAO's emotional reactivity (*PETS-ER*) was rated significantly higher in the experimental group (AV=29.1, SD= 5.72 vs AV=22.27, SD=6.13, $t(32) = 3.43$, $p=0.00$). Understanding and trust *PETS-UT* were also rated higher in the experimental group (AV=19.36, SD= 4.76 vs. AV=15.53, SD=4.64, $t(32) = 2.46$, $p = 0.0003$). Participants in the experimental group rated NAO higher on questions such as "NAO adapted to how I felt" and "NAO seemed capable of managing emotions", while similar responses between groups linked to a greater sense of trust as "NAO understood my goals" and "NAO inspired trust."

Finally, as far as *OE3* is concerned ("*Is there anything else you would like to add about your experience with NAO?*), participants in the experimental group provided particularly insightful feedback, emphasizing the robot's ability to convey empathy and enhance engagement. Several users noted specific interaction cues, such as the eye lighting and NAO's gaze-following behavior, as meaningful signs of responsiveness. Many described the experience as positive, highlighting NAO's ability to express and interpret emotions during the dialogue, and appreciating its friendly and effective communication style. These responses suggest that the empathic enhancements introduced in the experimental condition were noticed and appreciated, contributing to a more persuasive and emotionally engaging interaction. However, some critical reflections also emerged. One participant from the experimental group reported initial difficulty in relating to a robot, attributing this to the novelty of the experience, while another felt that the robot did not fully understand their responses. Notably, the latter participant's empathy score was relatively low (66.67%), which led to their exclusion from the persuasion analysis.

In the control group, feedback focused more on the robot's static behavior. Participants expressed a desire for more dynamic interaction, suggesting the inclusion of non-verbal cues to enrich the experience. These critiques, while valid, are consistent with the limitations of the control version, which lacked the empathic and expressive features present in the experimental condition.

## V. Discussion and Conclusion

This paper presented a controlled experiment. Participants were interviewed on their study habits by a social robot which either applied an emotional alignment strategy (experimental group) or neutral expressions (control group). We studied the effects of emotional alignment on the robot's persuasion potential (RQ1), participants' communication style (RQ2), particularly on emotivity and personal details disclosure, and participants' perception of the robot's emotional and cognitive capabilities (RQ3).

Our results show that emotionally aligned communication had limited effects on the robot's persuasiveness (H1). A significant difference in attitude after the emotionally aligned interview could be appreciated only once, regarding the statement on the benefit of caffeine consumption. Disparities were displayed in participants' communication style (H2, H3) only occasionally. These results are somehow unexpected and contrast with related work regarding communication style alignment [7], [9] and persuasion in both human-human [49] and human-computer interaction with non-embodied digital artefacts [35], [50], suggesting that different cognitive mechanisms might activate when humans interact with social robots instead of other humans or non-embodied digital artefacts. Actually, in the experimental condition, participants were significantly more prone to attribute mental states to the robot, considering both the epistemic, emotional and intentional dimensions (H4). Furthermore, NAO was perceived as more empathic regarding emotional responsiveness, understanding and trust dimensions, in comparison with control (H5). Hence, we conclude that the robot's emotional alignment was effectively conveyed but that it was unable to elicit significant mirroring effects in participants' communication style nor to facilitate persuasion. We surmise that either specific contextual factors, such as the chosen topic or the fact that the experiment took place in the very same environment where participants normally study, might have played a part.

Limitations of the study, as per persuasion evaluation, can be identified in the decision to measure variations in participants' disclosed agreement with the robot's attitudes and beliefs about the conversation topic without collecting data about participants' actual behaviours. Finally, participant feedback highlighted that NAO's empathetic responses were coherently transmitted, particularly through its use of eye colour changes and body movement. However, some participants initially found it challenging to interact with NAO. These results offer an intriguing clue to investigate further, with future studies concerning the effect of different multimodal capabilities enabling verbal and non-verbal emotional expressions, and how different communication modalities might serve different social robot's implementation goals.